%% file: latex/00_0_main.tex
\documentclass[11pt]{article}

\usepackage{acl}
\usepackage{amsmath}
\usepackage{fontspec}

\newfontfamily\hebrew{FreeSerif.otf}[
  Extension = .otf,
  Script=Hebrew
]
\newfontfamily\russian[Script=Cyrillic, Scale=1.0]{NotoSerif-Regular.ttf}
\usepackage{booktabs}
\usepackage[utf8]{inputenc}
\usepackage[T1]{fontenc}
\usepackage{float}
\usepackage{placeins}
\usepackage{bbm}
\usepackage{amsfonts}
\usepackage{times}
\usepackage{latexsym}
\usepackage{subcaption}
\usepackage{tabularx} 
\usepackage{booktabs}
\usepackage[T1]{fontenc}

\usepackage[utf8]{inputenc}

\usepackage{microtype}

\usepackage{inconsolata}

\usepackage{graphicx}

\title{A Universal Vibe? Finding and Controlling Language-Agnostic Informal Register with SAEs\thanks{~Accepted to EMNLP 2026.}}

\author{
  Uri Z. Kialy\thanks{~Equal contribution.} \\
  Ariel University \\
  \texttt{uri.kialy@ariel.ac.il} \\\And
  Avi Shtarkberg\footnotemark[2] \\
  Ariel University \\
  \texttt{avishb1213@gmail.com} \\\And
  Ayal Klein \\
  Ariel University \\
  \texttt{ayalk@ariel.ac.il} \\
}

\begin{document}
\maketitle

\input{latex/00_1_abstract}

\input{latex/01_intro}

\input{latex/02_bg}

\input{latex/03_dataset}

\input{latex/04_feature_identification}

\input{latex/05_observational}

\input{latex/06_casual_experiments}

\input{latex/07_discussion}

\input{latex/08_conclusion}

\input{latex/09_limitations}

\section*{Acknowledgments}
We used OpenAI's ChatGPT during manuscript preparation to assist with language editing, stylistic refinement, and the organization and phrasing of portions of the Introduction, Discussion, Conclusion, and other manuscript text. All generated suggestions were critically reviewed and revised by the authors, who take full responsibility for the final content.


\bibliography{custom}

\appendix

\input{latex/10_appendix}

\end{document}

%% file: latex/00_1_abstract.tex
\begin{abstract}
While multilingual language models successfully transfer factual and syntactic knowledge across languages, it remains unclear whether they process culture-specific pragmatic registers, such as slang, as isolated language-specific memorizations or as unified, abstract concepts. We study this by probing the internal representations of Gemma-2-9B-IT using Sparse Autoencoders (SAEs) across three typologically diverse source languages: English, Hebrew, and Russian. To isolate pragmatic register processing from trivial lexical sensitivity, we introduce a novel dataset in which every target term is polysemous, appearing in both literal and informal contexts. We find that while much of the informal-register signal is distributed across language-specific features, a small but highly robust cross-linguistic core consistently emerges. This shared core forms a geometrically coherent ``informal register subspace'', with greater geometric concentration at the deeper layer. Crucially, these shared representations are not merely correlational: activation steering with these features causally shifts output formality across all source languages and transfers zero-shot to six held-out languages spanning diverse language families and scripts. Together, these results provide the first mechanistic evidence that multilingual LLMs internalize informal register not just as surface-level heuristics, but as a portable, language-agnostic pragmatic abstraction.
\end{abstract}

%% file: latex/01_intro.tex
\section{Introduction}
\label{sec:intro}

While large language models (LLMs) have largely mastered multilingual syntax and factual recall, human communication is fundamentally driven by social signaling. When a speaker uses informal register or slang, they are not merely conveying information; they are negotiating social distance, signaling cultural belonging, and expressing casual intimacy. 
Because modern alignment pipelines overwhelmingly favor standard, highly edited text, LLMs suffer from a pervasive ``register gap.'' They frequently fail to appropriately process the dynamic, context-dependent language actually used in human interaction \citep{Sun2024TowardIL, li2024formality}, leading to miscalibrated responses and latent safety vulnerabilities where harmful intent is obfuscated by casual phrasing. Addressing this requires models calibrated not just for factual accuracy, but for deep sociolinguistic competence. 

This register gap exposes a fundamental, unresolved question about the architecture of multilingual representation: when an LLM encounters informal register across diverse cultures, does it process each instance as an isolated, language-specific surface form? Or does it construct a unified, language-agnostic abstraction of ``casual human interaction''? Prior mechanistic interpretability work has demonstrated that models build shared cross-lingual subspaces for facts and syntax \citep{conneau2020emerging, wendler_etal_2024_llamas}, but whether they abstract \emph{pragmatic} and \emph{social} concepts remains entirely unknown. We ask: \textit{Do LLMs represent informal register as a universal pragmatic concept, or as a disconnected collection of language-specific memorizations?}

To answer this, we probe the internal representations of Gemma-2-9B-IT using Sparse Autoencoders (SAEs). 
To isolate pragmatic register from the trivial confound of lexical bias (e.g., detecting informality merely due to the presence of specialized slang terms), we introduce a novel cross-lingual dataset across English, Russian, and Hebrew. Crucially, our dataset is strictly lexically controlled: every target term is polysemous, appearing in both its standard literal sense and its informal slang sense. This forces the model to resolve register from pragmatic context rather than mere vocabulary. 

What we find fundamentally challenges the view of models as mere surface-pattern matchers. By isolating latent SAE features that activate differentially across slang and literal contexts, we identify a small, geometrically coherent ``informal register subspace'' in the model's latent space, shared across all three typologically diverse source languages. 
Crucially, this abstraction is not merely correlational. Steering with this cross-linguistic feature set systematically and causally modulates output formality. Strikingly, this single sociolinguistic intervention transfers zero-shot to six typologically distant languages---including Japanese, Thai, Amharic, and Georgian---that were completely absent from our feature extraction pipeline. 
This zero-shot generalization provides compelling evidence that the model has internalized informal register not as a linguistic artifact, but as a shared, abstract pragmatic concept.

Our main contributions are threefold. 
\textbf{(i)~Methodological:} We introduce a novel, lexically-controlled dataset of polysemous terms across English, Hebrew, and Russian designed to isolate pragmatic register from any vocabulary bias. 
\textbf{(ii)~Observational:} We identify a geometrically coherent, language-agnostic ``informal register subspace'' within the SAE latent space, whose structure becomes more concentrated at greater model depth.
\textbf{(iii)~Causal:} We demonstrate that these shared latent features causally drive generation formality. The zero-shot transfer of this control to six typologically diverse, unseen languages provides the first mechanistic evidence for a cross-linguistic representation of pragmatic register.\footnote{Code available at
\url{https://github.com/UriKialy/universal_vibe.git}.}

%% file: latex/02_bg.tex
\section{Background}
\label{sec:bg}

\paragraph{The Register Gap in NLP}
While large language models exhibit remarkable multilingual fluency, their training and alignment are overwhelmingly anchored in standard, edited text. This creates a critical ``register gap'' when models encounter the dynamic, culturally situated nature of informal language \citep{li2024formality}. 
Prior multilingual research on register and style has largely focused on supervised formal-to-informal text transfer \citep{rao2018dear, briakou-etal-2021-ola, briakou-etal-2021-evaluating}. Multilingual resources for \emph{slang} remain exceptionally scarce; existing datasets are typically monolingual \citep{Sun2024TowardIL} or examine variation within individual languages rather than across diverse typologies \citep{wuraola-etal-2024-understanding}. 
Recent evaluations further show that even state-of-the-art models struggle to reliably deploy or comprehend informal register in context \citep{wu2025language}. 
Yet how multilingual models internally represent informal register---and whether these representations are shared across languages---remains largely unexplored.

\paragraph{Cross-Linguistic Model Interpretability}
A central debate in representation learning is whether multilingual models construct a shared, abstract ``interlingua'' or merely memorize parallel, language-specific surface patterns. Recent work has identified emergent cross-lingual alignment \citep{conneau2020emerging} and demonstrated that semantically equivalent concepts occupy shared geometric subspaces \citep{wendler_etal_2024_llamas}, often mediated by specialized multilingual neurons \citep{tang-etal-2024-language, liu-etal-2024-unraveling}. However, these investigations have heavily indexed on factual knowledge and syntactic structure. Much less is known about whether models develop abstract, shared circuitry for \emph{pragmatic register}---a dimension of language that is inherently social and culturally idiosyncratic. While prior work has successfully identified monolingual steering directions and style-sensitive neurons \citep{subramani2022extracting, rimsky2024steering, lai2024style}, it remains an open question whether these mechanisms transfer cross-lingually. Our results bridge this gap, examining the balance between shared and language-specific latent features to determine if models possess a universal pragmatic architecture.

\paragraph{Sparse Autoencoders and Activation Steering}
To untangle these complex pragmatic representations, we utilize Sparse Autoencoders (SAEs). By decomposing dense neural activations into an overcomplete, sparse basis, SAEs effectively mitigate polysemanticity, revealing human-interpretable features that standard neuron-level analyses obscure \citep{bricken2023towards, templeton2024scaling, cunningham2023sparse}. Beyond serving as observational lenses, SAEs unlock \emph{activation steering}: the ability to intervene on specific latent features at inference time to test their causal role in model behavior. 
This technique has successfully modulated model safety, mitigating social biases \citep{durmus2024evaluating} and controlling refusal behaviors \citep{obrien2024steering, arad2025saes}. We extend this causal framework to the sociolinguistic domain, using SAE-derived steering vectors to test whether a shared cross-linguistic ``slang subspace'' can predictably control output register across diverse languages.

%% file: latex/03_dataset.tex
\section{Dataset}
\label{sec:datasets}

To evaluate cross-lingual register representations without the confound of lexical bias, we constructed a novel dataset across English, Hebrew, and Russian. To our knowledge, this is the first slang benchmark spanning more than two typologically distinct languages \citep{mei_etal_2024_slang, hislang2024, wuraola-etal-2024-understanding}. 

Our core design principle is strict lexical control. 
Prior to sentence collection, native speakers manually compiled a set of \textit{polysemous} target terms, each carrying both a standard literal sense and a low-register informal sense (see Table \ref{tab:dataset_examples}). 
By evaluating register classification on sentences sharing the exact same target token, we ensure models cannot rely on word identity, out-of-vocabulary novelty, or term frequency; instead, they must resolve the register entirely from pragmatic context.

Instance-level labels (\textit{Slang} vs.\ \textit{Literal}) were assigned using GPT-4o-mini \citep{openai2024gpt4o}. While prior work establishes the broader LLM-as-a-judge paradigm for text evaluation \citep{zheng2023judging, pangakis2023automated, gilardi2023chatgpt}, the specific reliability of this framework for formal/informal register distinction is validated by our own human evaluation. Corroborated by author manual review, the automated labeling achieved a precision of $\geq 0.92$ across all three languages (detailed in Appendix~\ref{app:label_validation}). Detailed dataset statistics and slang-to-literal ratios are provided in Appendix~\ref{app:dataset_stats}.\footnote{For access to the dataset,
please contact the authors.}

\begin{table}[t]
\centering
\small
\begin{tabular}{lll}
\toprule
\textbf{Term} & \textbf{Usage} & \textbf{Example Sentence} \\
\midrule
\textit{fire} & Literal & It says there was a big fire in \\
              &         & London 300 years ago. \\
\midrule              
\textit{fire} & Slang   & That movie was straight fire. \\
\midrule
\textit{sick} & Literal & I think I better go home \\
              &         & before I'm sick. \\
\midrule              
\textit{sick} & Slang   & My car's got fresh 20's, \\
              &         & they look sick! \\
\bottomrule
\end{tabular}
\caption{Examples of polysemous target terms. The identical token forces register disambiguation from context alone.}
\label{tab:dataset_examples}
\end{table}

\subsection{Corpus Composition}
\label{Corpus Composition}
\textbf{English.} Comprises 2,835 labeled instances spanning 130 target terms (e.g., \textit{fire}, \textit{wet}, \textit{sick}). Sentences were sourced from curated HuggingFace corpora and the naturalistic colloquial corpus introduced by \citet{Sun2024TowardIL}.

\textbf{Hebrew.} Contains 4,527 labeled instances across 18 target terms
(e.g.\ {\hebrew אש} -- ``fire/awesome''). Data was sourced from HuggingFace
corpora (theatrical text, forums, lyrics). Hebrew's non-concatenative
morphology makes our strictly controlled token constraint particularly vital,
as inflected surface forms of slang vary substantially.

\textbf{Russian.} Consists of 1,259 labeled instances covering 15 target terms (e.g., {\russian пушка} --``gun/banger''). Sentences were organically sourced via API from VKontakte (VK) and Telegram, providing an ecologically valid representation of contemporary informal Russian.

\textbf{Lexical diversity.} Lexical concentration varies substantially across languages, with a broader target-term distribution in English and greater concentration in Hebrew and Russian (Appendix~\ref{app:dataset_stats}).

%% file: latex/04_feature_identification.tex
\section{Latent Feature Extraction and Analysis Setup}
\label{sec:feature_identification}

We use Gemma-2-9B-IT, leveraging its extensive suite of publicly available 131K-feature Sparse Autoencoders \citep{lieberum2024gemma}. Preliminary inspection identified substantial feature activity at Layer 9 and Layer 20, which serve as our mid- and deep-layer comparative baselines, respectively.\footnote{A deeper layer (Layer 31) yielded negligible target-term activity passing our minimum threshold---likely because final layers heavily compress representations toward next-token prediction---and was therefore excluded from analysis.}

\subsection{SAE Feature Extraction}
\label{sec:sae_feature_extraction}

To isolate contextualized register representations from trivial lexical confounds, our extraction pipeline focuses strictly on the \emph{target term} (the polysemous word used in either a slang or literal sense). 
For each sentence, we extract the model's residual stream activations at the subword positions spanned by the target term and encode each activation via the pretrained SAE into a sparse feature vector:
\[
\mathbf{f} = (f_1, \dots, f_{131072}), \qquad f_i \ge 0,
\]
where each dimension corresponds to one latent SAE feature. We apply this procedure independently across all three source languages to systematically evaluate the cross-lingual intersection of these target-term representations.

\subsection{Feature Scoring and Selection}
\label{sec:feature_scoring}
To identify slang-associated latent features, we compute the \textit{differential activation frequency} for each feature $i$:
\begin{equation}
\Delta_i = P(f_i > 0 \mid \text{slang}) - P(f_i > 0 \mid \text{literal}).
\label{eq:delta}
\end{equation}
We estimate these probabilities independently for each language and layer by pooling all target-term subword positions corpus-wide. 
Each subword position is treated as a separate observation and inherits the slang/literal label of its corresponding target-term instance; consequently, targets split into multiple subwords contribute multiple observations. 
Thus, $\Delta_i$ measures how much more frequently feature $i$ activates at target-term positions in slang than in literal contexts.
For subsequent analyses and experiments, we rank all 131{,}072 SAE features by $\Delta_i$ and retain the top-$k$ features, using $k{=}100$ for defining the cross-linguistic overlap set (\S\ref{sec:derived_feature_sets}).

Separately, for the pooled feature-quality analysis in \S\ref{subsec:feature_distr}---where each feature is evaluated as a standalone binary classifier and rare features would yield unstable precision and recall estimates---we apply a \textit{minimum activity filter} consisting of two criteria: a feature must activate on at least 5\% of slang tokens and fire a minimum of 10 times across the combined dataset.
These dual thresholds ensure that retained features represent a meaningful fraction of informal usage---rather than isolated idiosyncratic phrases---while accumulating sufficient absolute frequency for stable statistical estimation.

We stress that the two steps are independent and applied in this order: the per-language top-$k$ rankings are computed on the unfiltered $\Delta_i$ ordering, and the activity filter is applied only to the pooled classifier analysis.
Applying the filter before ranking would alter which features enter the top-100 lists, and hence the composition and size of the cross-linguistic core reported in \S\ref{sec:derived_feature_sets}.

\subsection{Derived Feature Sets}
\label{sec:derived_feature_sets}

Intersecting the top-100 ranked feature lists across the three languages, we identify a \textit{cross-linguistic} feature core: features appearing in the top-100 lists of all three languages simultaneously---9 features at Layer~9 and 10 at Layer~20.
We also investigate high-salience bilingual feature sets for completeness (see details in Appendix ~\ref{app:bilingual}), but our primary focus is on the cross-linguistic core as the strongest candidate of shared informal register representation.

These set definitions play two roles in the paper. First, they support the
observational analyses in Section~\ref{sec:observed_findings}, where we ask
whether slang-related features exhibit shared cross-linguistic structure.
Second, they seed the feature pools used later for activation steering
(Section~\ref{sec:causal_experiments}), where we test whether intervening on these feature
sets causally shifts output register.

\FloatBarrier

%% file: latex/05_observational.tex
\section{Observational Findings}
\label{sec:observed_findings}

We evaluate the extracted feature sets across four dimensions: individual discriminative quality, cross-lingual overlap, geometric coherence, and lexical readout. 
Across both analyzed layers, a consistent picture emerges: while the informal-register signal is distributed over multiple latent features, a small but non-trivial subset of these features is shared across languages, and this shared subset becomes more geometrically coherent at the deeper layer.

\subsection{Individual Feature Quality and Distributed Signal}
\label{subsec:feature_distr}

We begin by evaluating whether slang detection is dominated by a few especially strong latent features or instead distributed across a broader set. To do so, we treat each of the 131{,}072 SAE features as a binary classifier on the pooled cross-linguistic dataset (5,256 slang subword tokens; 7,697 literal subword tokens), using the rule \emph{active} $\Rightarrow$ \emph{slang}, and retaining only features that pass the minimum activity filter from Section~\ref{sec:feature_scoring}.
These counts are at the subword-token level rather than the instance level: the extraction pipeline (\S\ref{sec:sae_feature_extraction}) records the residual-stream activation at every subword token spanned by a target term, so each labeled instance contributes one or more tokens depending on Gemma-2's tokenizer. The ratio is close to $1.0$ for English (977 tokens from 968 slang instances) but approaches $1.9$ for Hebrew and Russian, whose target terms are frequently split into multiple subwords; instance-level counts are given in Appendix~\ref{app:dataset_stats}, Table~\ref{tab:dataset_stats}.

The resulting feature landscape shows a clear precision--recall trade-off at
both layers. Some features are relatively precise but fire on only a limited
subset of slang tokens, whereas others provide broad coverage at lower precision. For example, high-$\Delta_i$ features such as 35440 (Layer~20,
precision 0.849) and 38266 (Layer~9, precision 0.776) achieve relatively clean slang selectivity with moderate recall, while broad-coverage features such as 127994 (Layer~9) and 26379 (Layer~20) capture more than 85\% of slang tokens at substantially lower precision.
No single feature dominates all classifier metrics: Feature~93521 at Layer~20 is the only feature that appears simultaneously in the top-10 lists for precision, recall, and F1 (precision 0.966, recall 0.240, F1 0.384), but it leads none of them individually.

This distributed pattern motivates our focus on \emph{feature sets} rather than
individual features. The latent representation of informal usage is not
concentrated in a single ``slang neuron'', but spread across multiple partially
overlapping sparse features.

\subsection{Cross-Linguistic Feature Overlap}
\label{sec:cross_linguistic_overlap}
We next ask whether the most slang-associated features are largely
language-specific or whether some are shared across languages. Intersecting the
top-100 $\Delta_i$-ranked features for English, Hebrew, and Russian at each
layer yields a consistent three-tier structure: a small cross-linguistic core,
a moderate bilingual set, and a larger language-specific remainder.
The cross-linguistic core is small at both analyzed layers---9 features at
Layer~9 and 10 at Layer~20---against 32 and 30 features respectively appearing
in the top-100 lists of exactly two languages, and 209 and 210 appearing in
exactly one. Thus, most slang-related latent features are not cross-linguistic,
but the model consistently contains a small shared subset that survives across
all three languages.
Among language pairs, Hebrew--Russian shares more exclusive features than
Hebrew--English at both layers (15 vs.\ 6 at Layer~9; 14 vs.\ 5 at Layer~20),
possibly reflecting shared lower-resource status in the training distribution;
English--Russian accounts for the remaining 11 at each layer.
\paragraph{Statistical Significance of Overlap Counts.}
To test whether this overlap is greater than expected by chance, we perform a
label-permutation test ($n=10{,}000$).
In each iteration, we independently shuffle slang/literal labels within each
language, recompute feature rankings by $\Delta_i$, and count the resulting
triple overlap among the top-100 lists.
This permutation preserves the marginal activation frequencies and sparsity
dynamics of every feature, isolating the target effect by destroying only the
semantic mapping between activations and the slang--literal pragmatic labels.
Under this null, the mean triple overlap is 0.74 (std $= 0.90$), and the maximum
observed across all permutations is 7.
The observed overlaps of 9 (Layer~9) and 10 (Layer~20) therefore lie outside the
entire null distribution ($p < 10^{-4}$), supporting the conclusion that the
cross-linguistic core reflects genuine shared structure tied to the
slang--literal contrast rather than incidental feature co-occurrence.
Taken together, these results suggest that the model represents informal usage
through a mixed architecture: predominantly language-specific and bilingual
features, plus a small but statistically robust cross-linguistic core.

\subsection{Geometric Coherence of the Shared Feature Set}
\label{sec:geometric}
\begin{figure}[!ht]
    \centering
    \includegraphics[width=\columnwidth]{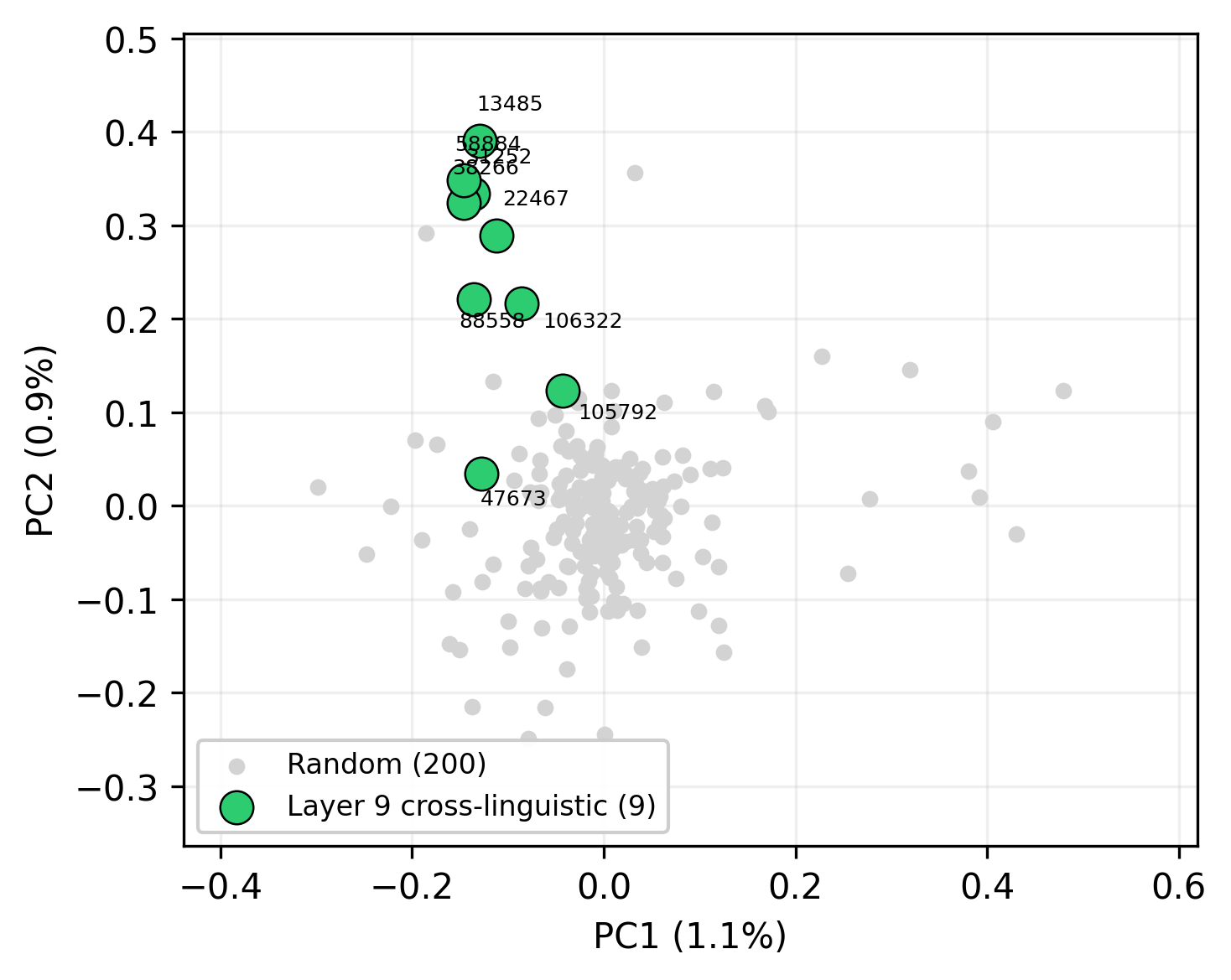}
    \small (a) Layer 9 (island score 4.3$\times$)
    \vspace{0.3cm}
    \includegraphics[width=\columnwidth]{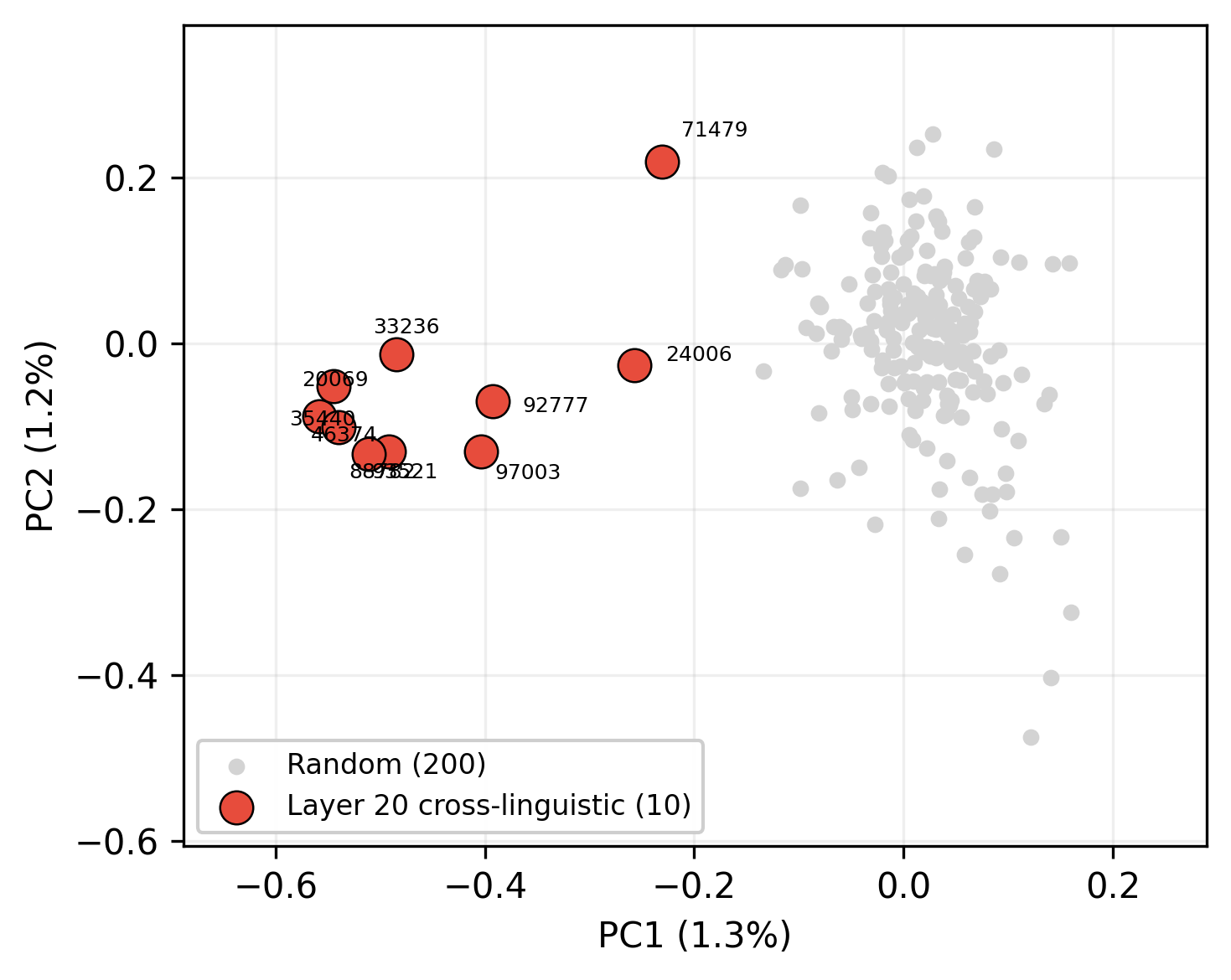}
    \small (b) Layer 20 (island score 7.8$\times$)
    \caption{Illustrative projection of cross-linguistic informal-register feature decoder vectors (colored) against randomly sampled SAE features (grey). Cross-linguistic features form a tighter grouping at Layer~20 than at Layer~9. The quantitative evidence comes from cosine-similarity analysis in the full decoder space; the 2D projection is shown only as a visual illustration, using an independent 200-feature random subsample.}
    \label{fig:pca_geometry}
\end{figure}

Feature overlap alone does not imply that the shared features form a coherent
latent subspace: in principle, the same features could recur across languages
without occupying a meaningful region of decoder space. We therefore examine the
geometry of the cross-linguistic feature core using the SAE decoder vectors
$\{\mathbf{w}_i^{\text{dec}}\}_{i \in \mathcal{F}_{\text{cross}}}$.
For each layer, we compute the mean pairwise cosine similarity among the decoder
vectors of the cross-linguistic features, and compare it to the mean
\emph{absolute} cosine similarity between those features and 1{,}000 randomly
sampled SAE features, averaged over five independent random draws. We take
absolute values because the signed core-to-random mean is essentially zero at
both layers ($+0.0004$ at Layer~9, $-0.0008$ at Layer~20); the magnitude of the
alignment, not its sign, is the meaningful reference scale. The random sample
itself is internally unstructured, with near-zero mean pairwise cosine
similarity ($0.0007$ at Layer~9 and $0.0014$ at Layer~20), indicating that any
observed structure is not simply an artifact of the global SAE geometry.
By contrast, the cross-linguistic features exhibit substantially elevated
within-group similarity. To quantify this clustering, we define an
\textit{island score} as the ratio of the mean pairwise cosine similarity within
the cross-linguistic set to its mean absolute cosine similarity with the random
sample. A score substantially above $1\times$ indicates a geometrically coherent
cluster. At Layer~9, the mean within-group similarity is $0.065$ against a
random reference of $0.0150$, an island score of 4.3$\times$. At Layer~20,
coherence sharpens to $0.152$ against $0.0195$, a 7.8$\times$ island score. Both
values are stable under resampling (4.32--4.38 and 7.82--7.95 across five random
seeds). Thus, the cross-linguistic features do not merely overlap by rank; they
occupy a denser and more coherent region of decoder space, especially at the
deeper layer (Figure~\ref{fig:pca_geometry}).
This geometric result strengthens the overlap analysis: the cross-linguistic
core is not just statistically non-random, but also internally structured in a way consistent with a shared latent direction family.

\subsection{Lexical Readout via Vocabulary Projection}
\label{sec:vocab_projection}
As a post-hoc interpretability probe, we project each cross-linguistic feature's
decoder weight vector through the model's \emph{unembedding matrix}
$W_U \in \mathbb{R}^{d \times |V|}$ ---the model's final linear map from a
$d$-dimensional residual-stream vector to logits over the vocabulary $V$, and the
output-side counterpart of the input embedding (tied to it in Gemma-2-9B-IT)---%
following \citet{gur-arieh-etal-2025-enhancing}, in order to inspect which
vocabulary items each latent direction most promotes.
Roughly half of the core features yield token lists clearly enriched for informal
or register-marked expressions---including \textit{dudes}, \textit{vibes},
\textit{swag}, and \textit{dope}, alongside evaluative intensifiers and
profanity---while the remainder promote tokens with no readily interpretable
lexical structure. We therefore interpret this not as a standalone proof of
feature meaning, but as a qualitative sanity check that a substantial subset of
the identified cross-linguistic features have lexical readouts consistent with
informal register. Examples for the two highest-$\Delta_i$ features are provided in
Appendix~\ref{app:vocab_proj}.
Interestingly, features whose discriminative signal derives from Hebrew or
Russian alone---including ones with no English slang--literal signal at all---
often still project to English tokens, suggesting a systematic English bias in
the \emph{unembedding space} (the column space of $W_U$, spanned by the
per-token output directions). This bias manifests at generation time as
occasional language switches toward English, though these affect approximately
10\% of steered completions overall (Appendix~\ref{app:lang_preservation}). For
this reason, we use vocabulary projection only as a supporting interpretability
analysis: it is informative about the lexical tendencies of a latent direction,
but it does not by itself establish the language-specificity or cross-linguistic
generality of that feature. Those properties are instead supported by the overlap
and steering analyses.

\subsection{A Descriptive Depth Trend}
\label{subsec:depth}
Although our main results focus on the cross-linguistic feature core, the comparison between Layer 9 and Layer 20 reveals a consistent descriptive trend.

First, the model represents informal register far more sparsely at the deeper layer.
Because features at Layer 20 are highly specialized, fewer of them activate frequently enough to pass our 5\% minimum activity threshold---the requirement that a feature fire on at least 5\% of slang tokens---with 169 pooled features passing at Layer~20 against 892 at Layer~9.
The same contrast appears at the token level: at Layer~9, slang targets trigger broader activation than literal targets, whereas at Layer~20 they engage fewer active features at lower total activation magnitude.
Second, the cross-linguistic core identified at Layer~20 is more geometrically coherent than its Layer~9 counterpart, as reflected in its higher within-group cosine similarity and larger island score.

Taken together, these observations suggest a depth-dependent reorganization of informal-register information: the signal is more broadly distributed at Layer 9, while at Layer 20 it is represented by a sparser set of features occupying a more geometrically coherent region of decoder space.
We treat this as a descriptive empirical trend rather than a strong mechanistic claim.

%% file: latex/06_casual_experiments.tex
\raggedbottom

\section{Causal Experiments}
\label{sec:causal_experiments}
We next test whether the cross-linguistic informal-register SAE features identified above can causally control register during generation, rather than merely tracking it as a correlational artifact. We use the Layer 20 cross-linguistic core as our primary intervention set and report complementary Layer 9 experiments in Appendix~\ref{app:formality_by_language}. Crucially, testing these interventions across linguistic boundaries allows us to assess the extent to which these features encode a cross-linguistic, abstract representation of pragmatic register.

\subsection{Experimental Setup}

\paragraph{Steering Mechanism}
\label{subsec:Steering_Mechanism}
For each feature set $\mathcal{F}$ (e.g., cross-linguistic features), we construct a 
steering vector $\mathbf{v}_{\mathcal{F}}$ by averaging the SAE decoder weight vectors 
of its constituent features:
\begin{equation}
\mathbf{v}_{\mathcal{F}} = \frac{1}{|\mathcal{F}|}\sum_{i \in \mathcal{F}} \mathbf{w}_i^{\text{dec}}
\label{eq:steering_vector}
\end{equation}
where $\mathbf{w}_i^{\text{dec}}$ is the decoder weight vector for feature $i$. The 
resulting vector is L2-normalized: $\mathbf{v}_{\mathcal{F}} \leftarrow 
\mathbf{v}_{\mathcal{F}} / \|\mathbf{v}_{\mathcal{F}}\|_2$. During generation, we 
inject the steering vector into the post-block residual stream at layer $L$ via a 
forward hook applied at every generated token position $t$:
\begin{equation}
\mathbf{h}'_{L,t} = \mathbf{h}_{L,t} + \alpha \cdot \mathbf{v}_{\mathcal{F}}
\end{equation}
where $\mathbf{h}_{L,t}$ is the residual stream activation at layer $L$ and token 
position $t$, $\alpha$ is the steering coefficient, and $\mathbf{h}'_{L,t}$ is the 
modified activation. Negative $\alpha$ steers toward formal language; positive $\alpha$ 
steers toward informal language. We test 6 coefficients ranging from $-150$ to $100$.
We construct steering vectors from the cross-linguistic feature set: the 10-feature core at Layer~20 for the primary experiment (\S \ref{subsec:causal_results}), and the  9-feature core at Layer~9 (Appendix \ref{app:formality_by_language}).
Results for bilingual-exclusive 
feature sets are reported in Appendix~\ref{app:bilingual}.

\paragraph{Prompts and Languages}
We use 20 open-ended sentence prompts (e.g., ``I think that\ldots'') across 
nine languages (see Appendix~\ref{app:steering_prompts} for the full prompt 
list), yielding ${\sim}1{,}080$ sentence completions. 
The prompts were translated from English into Hebrew, Russian, and German by native speakers, with translations validated by GPT-4o, and into the remaining five languages by the same model directly.
Our evaluation spans three \emph{source} languages whose data contributed 
to feature extraction (English, Hebrew, and Russian) and six \emph{zero-shot 
generalization} languages entirely absent from our feature extraction data: 
German, Japanese, Hindi, Thai, Georgian, and Amharic. The six zero-shot 
languages were chosen to maximize typological diversity, spanning six 
different language families (Germanic, Japonic, Indo-Aryan, Kra-Dai, 
Kartvelian, and Semitic) and six distinct script traditions.

\paragraph{Evaluation}
Formality is evaluated on a continuous scale $[0, 1]$ using GPT-4o-mini as an automated judge, where 1 represents highly formal and 0 highly informal language. Steering effectiveness is measured via Pearson correlation between $\alpha$ and formality score, computed over all individual completions (i.e., 120 per language per feature set). 
To validate the automated formality scores, we manually rated 30 completions per language for four languages: the three source languages (English, Hebrew, and Russian) and one zero-shot language (German). 
For each language, the completions were scored by one fluent speaker of that language, drawn from the authors or their fluent acquaintances. 
GPT-4o-mini scores correlated with the human ratings at $r = 0.62$, indicating moderate agreement. Given the inherently subjective nature of formality judgments, we treat the automated scores for the remaining five zero-shot languages as reasonable proxies for human judgment.

\subsection{Results}
\label{subsec:causal_results}

\input{tables/steering_results}

\paragraph{Source Language Control and Bilingual Transfer}
As detailed in Table~\ref{tab:steering_summary}, applying the cross-lingual steering vectors to the three source languages produces clear, statistically significant negative correlations between the steering coefficient $\alpha$ and output formality ($r$ between $-0.40$ and $-0.51$, all $p < 10^{-5}$). This confirms that the shared feature set causally modulates register within its derived languages.

As an intermediate test of cross-lingual abstraction, we evaluate \emph{bilingual-exclusive} feature sets---defined from two source languages and applied exclusively to the held-out third. These feature sets also produce significant monotonic decreases in formality (Appendix~\ref{app:bilingual}), establishing that causal register control does not require the target language itself to contribute directly to feature extraction.

\begin{figure}[t]
    \centering
    \includegraphics[width=\linewidth]{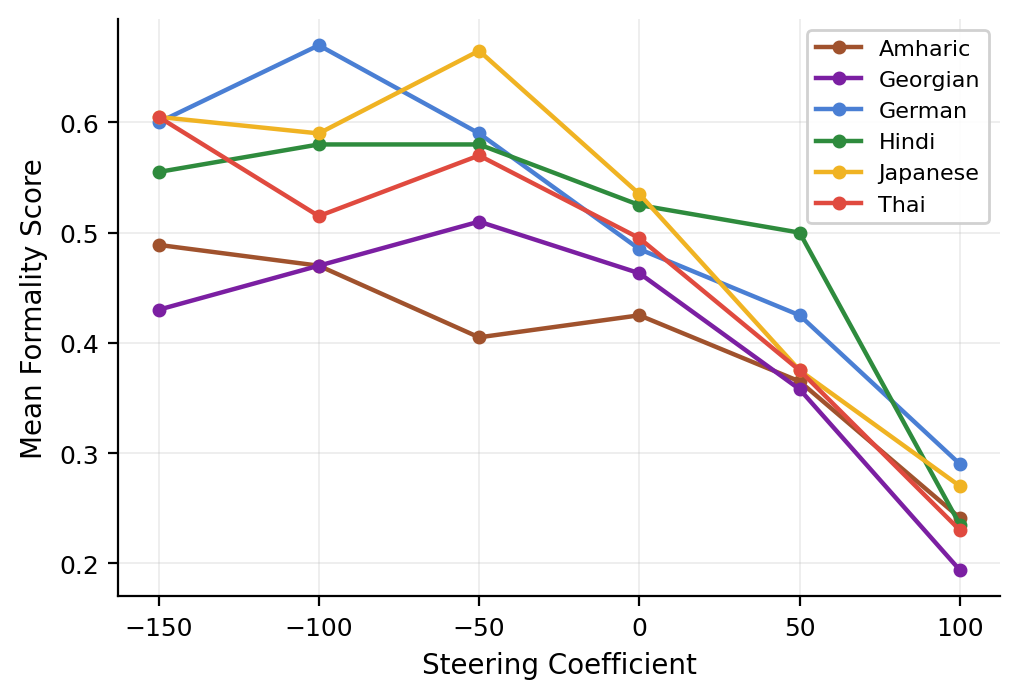}
    \caption{Per-language formality scores under cross-lingual feature steering for all six zero-shot transfer languages. All zero-shot languages exhibit a formality decrease as steering coefficient $\alpha$ increases, with the sharpest changes in the informal direction.}
    \label{fig:zero_shot_formality}
\end{figure}

\paragraph{Zero-Shot Transfer to Unseen Languages}
Building on this abstraction, we test whether this causal control transfers zero-shot to six languages entirely absent from feature identification. Despite their typological distance from the source-language set, every tested language shows a significant decrease in formality as $\alpha$ increases, with the effect concentrated in the informal direction (Figure~\ref{fig:zero_shot_formality}).

The strongest zero-shot transfer occurs in Japanese and Thai, while even Georgian and Amharic exhibit moderate but highly significant negative correlations ($p<10^{-3}$). Across the zero-shot set, mean formality decreases substantially (from $0.548$ at $\alpha=-150$ to $0.245$ at $\alpha=100$), demonstrating that the cross-lingual features successfully modulate register in typologically diverse, zero-shot settings. 

Complementary Layer 9 experiments reproduce the same qualitative steering pattern across all three source languages, with correlations somewhat larger in magnitude than those observed at Layer 20 (Appendix~\ref{app:formality_by_language}).\footnote{This contrasts with \S \ref{sec:observed_findings}, where the Layer 20 cross-linguistic features are more geometrically coherent than those at Layer 9. This dissociation suggests that geometric organization and steering efficacy capture distinct aspects of the representation rather than necessarily increasing together.}
Furthermore, comparison against a random feature baseline confirms that these formality shifts are specific to the identified cross-lingual subspace rather than a generic property of residual-stream perturbation (Appendix~\ref{app:random_ablation}).

\begin{figure}[H]
    \centering
    \includegraphics[width=\linewidth]{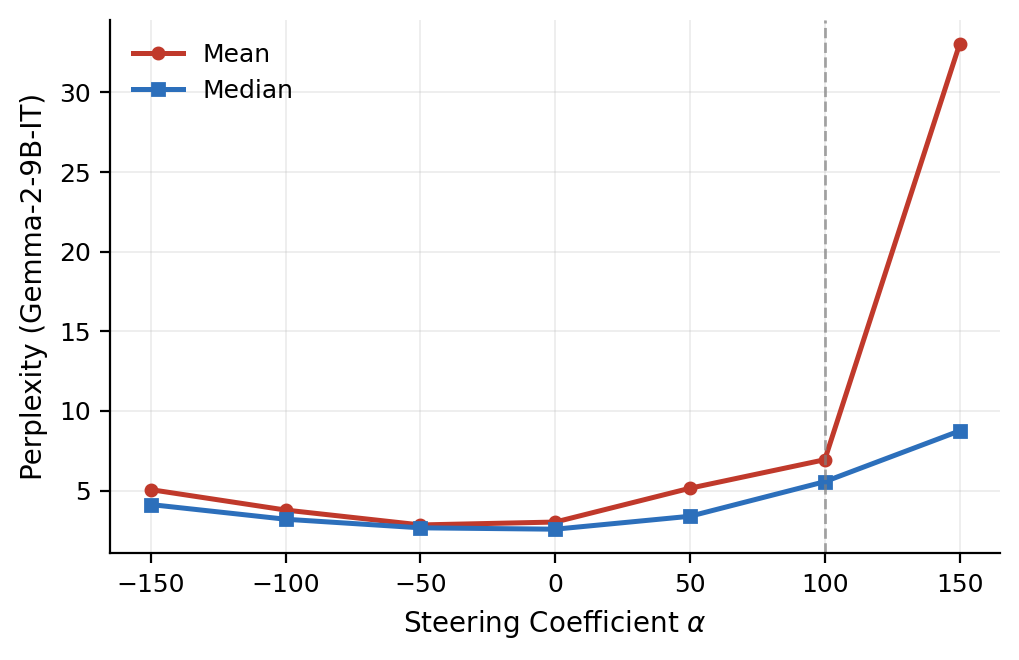}
    \caption{Mean and median perplexity (Gemma-2-9B-IT self-perplexity) across
    all nine languages as a function of steering coefficient $\alpha$.
    Perplexity is stable across $[-150, 100]$ and rises sharply beyond
    $\alpha = 100$, motivating the chosen operating range.}
    \label{fig:perplexity_gemma}
\end{figure}

\paragraph{Native Slang, Not Translation}
Beyond aggregate formality scores, we test whether steered completions remain
coherent and in the target language. Gemma-2-9B-IT self-perplexity remains low
and stable across all nine languages throughout $\alpha \in [-150, 100]$, then
rises sharply beyond $\alpha = 100$ (Figure~\ref{fig:perplexity_gemma}),
motivating this interval as our operating range. External reference models
(Qwen2.5-7B-Instruct, and DictaLM 2.0 for Hebrew) show the same pattern
(Appendix~\ref{app:perplexity}). Automatic language identification further
shows that approximately 90\% of completions remain in the target language,
with most switches directed toward English
(Appendix~\ref{app:lang_preservation}).

Qualitative manual inspection further shows that informal steering can be realized through language-specific slang rather than direct translations of English expressions. 
For the shared prompt stem ``In my opinion, the movie was\ldots'', strong formal steering ($\alpha{=}{-}150$) yields measured, explanatory prose in every language, whereas strong informal steering ($\alpha{=}{+}100$) produces casual completions with language-specific informal expressions:
English \emph{so good}, Russian {\russian крутым} (``awesome'') and {\russian бомба} (``a blast''), Hebrew {\hebrew ממש כיף} (``really fun''). 
These distinct realizations are consistent with steering a shared pragmatic representation rather than reproducing a fixed English lexical pattern.
Steering is similarly effective on English stems that begin in an informal
register. 
Formal steering still strips out the casual tone---``Yo, that party
was\ldots'' becomes ``a lot of fun. I met some interesting people''---while
informal steering intensifies it further, as in ``\ldots \textbf{LIT} last
night!''

%% file: tables/steering_results.tex
\begin{table}[t]
\centering
\small
\setlength{\tabcolsep}{12pt}
\begin{tabular}{lc}
\toprule
\textbf{Target Language} & \textbf{$r$ (L20)} \\
\midrule
English & $-0.506$ \\
Hebrew  & $-0.398$ \\
Russian & $-0.505$ \\
\midrule
 German   & $-0.458$ \\
 Japanese & $-0.508$ \\
 Hindi    & $-0.425$ \\
 Thai     & $-0.499$ \\
 Georgian & $-0.328$ \\
 Amharic  & $-0.437$ \\
\bottomrule
\end{tabular}
\caption{Summary of steering results --- Pearson $r$ correlation between $\alpha$ (steering strength) and generated text formality, computed over the operating range $\alpha \in [-150, 100]$. The cross-linguistic feature set effectively controls the register of generated texts at all source languages, and this transfers to six zero-shot languages 
absent from the feature extraction pipeline. Results of steering in Layer 9 are reported in  Appendix~\ref{app:formality_by_language}.} 
\label{tab:steering_summary}
\end{table}

%% file: latex/07_discussion.tex
\section{Discussion}
\label{sec:discussion}

\paragraph{What Does the Shared Subspace Represent?}
Our findings may reflect a representation broader than slang alone. The 9--10
cross-linguistic features we identify may partly capture a broader social-pragmatic
function of informal language---for example, in-group signaling through
linguistic deviation---that recurs across languages
\citep{greenberg1963universals, eble1996slang, mattiello2008introduction}.
This interpretation is consistent with the transfer of steering to languages
absent from feature identification, and predicts that related phenomena such
as code-switching, taboo language, or in-group jargon may engage overlapping
cross-linguistic features. In this sense, the ``slang island'' may provide a
window into a broader pragmatic representational space.

At the same time, such a representation should not be expected to separate cleanly from semantics. Our contrastive design controls for target-word identity by comparing literal and slang uses of the same lexical item, but these uses can still differ in meaning, topic, affect, and social stance.
Indeed, strict semantic--pragmatic disentanglement seems to be conceptually ill-posed: register is itself realized partly through semantic and affective choices. We therefore interpret the identified subspace as strongly associated with informal register rather than as a fully content-abstract register representation. 
Its cross-linguistic recurrence, causal steerability, and register-marked lexical readout provide converging evidence for this interpretation, while more tightly controlled contrasts could further characterize what information the subspace contains.

\paragraph{From Features to Circuits.}
The identified SAE features also provide a natural starting point for
circuit-level analysis \citep{elhage2021mathematical}. Activation patching
\citep{wang2023interpretability} and emerging SAE feature-circuit methods
\citep{marks2025sparse} could test whether these features participate in a
shared causal pathway for register processing across languages, extending the
present feature-level analysis toward a mechanistic account of pragmatic
control.

%% file: latex/08_conclusion.tex
\section{Conclusion}
\label{sec:conclusion}

We investigated whether multilingual LLMs represent informal register through language-specific features alone or also through shared latent structure.
Using Sparse Autoencoders in Gemma-2-9B-IT, we identified a small, geometrically coherent cross-linguistic feature core shared across English, Hebrew, and Russian. 
Steering this feature set causally modulates output formality and transfers zero-shot to six languages absent from feature identification.
Together, these findings provide evidence that multilingual LLMs encode informal register through a partly shared, abstract pragmatic representation rather than solely through language-specific surface patterns.

%% file: latex/09_limitations.tex
\section{Limitations}
\label{sec:limitations}

While our findings provide mechanistic evidence for cross-lingual pragmatic representations, several limitations contextualize our results. 
First, our layer selection was constrained by the availability of pre-trained SAEs; a comprehensive layer sweep could reveal a more precise depth profile of slang representation. 
Furthermore, our experiments are conducted exclusively on Gemma-2-9B-IT; whether this geometrically coherent ``informal register island'' generalizes across other model architectures or scale regimes remains an open question.

Second, we operationalize informal register primarily through slang-sensitivity (the slang--literal contrast) evaluated at the target-term level. Sociolinguistically, slang (ephemeral, in-group vocabulary) and informal register (a broad, stable communicative mode) are distinct, and register shifts often manifest as sequence-level syntactic relaxations rather than isolated lexical choices. 
However, our causal steering experiments demonstrate that intervening on these token-derived features systematically controls broader generation formality. 
Furthermore, the steering results extend beyond the source languages and are not reducible to target-word identity alone, supporting a broader pragmatic interpretation of these features.

Finally, our evaluation methodology and the model's own latent geometry exhibit an English-centric bias. 
Vocabulary projection of the universal features heavily favors English terms, and steered generation occasionally induces language switching toward English, particularly at stronger steering coefficients ($\alpha > 100$). 
This suggests that the "language-agnostic" subspace may function partially as an English-anchored hub rather than a perfectly symmetric abstract interlingua.
Additionally, relying on GPT-4o-mini as an automated formality judge for typologically distant languages (e.g., Amharic, Thai) risks projecting Western-aligned definitions of formality onto cross-cultural pragmatics. 
Future work incorporating comprehensive native-speaker evaluations will be necessary to fully disentangle universal pragmatics from pretraining data imbalances.

%% file: latex/10_appendix.tex

\section{Dataset Statistics}
\label{app:dataset_stats}
\begin{table}[H]
\centering
\small
\begin{tabular}{lrrrr}
\toprule
 & English & Hebrew & Russian & Total \\
\midrule
Labeled instances     & 2,835 & 4,527 & 1,259 & 8,621 \\
Slang instances       &   968 & 1,776 &   538 & 3,282 \\
Literal instances     & 1,857 & 2,751 &   721 & 5,329 \\
Unique target terms   &   130 &    18 &    15 & --- \\
\bottomrule
\end{tabular}
\caption{Dataset statistics per language. Counts are labeled
(sentence, target-term) instances: a sentence containing two target terms
contributes two instances. Ten English instances carry an inconclusive judge
label and are excluded from both classes, so the slang and literal rows sum to
8,611 rather than 8,621.}
\label{tab:dataset_stats}
\end{table}

This section provides a detailed breakdown of the multilingual dataset used in our experiments. Table \ref{tab:dataset_stats} summarizes the distribution of slang and literal instances across the three source languages: English, Hebrew, and Russian.
We collect sentences via two complementary methods. First, we sample from existing slang corpora such as OpenSubtitles-Slang ~\citep{Sun2024TowardIL}, retaining only sentences where a target term from our human-curated list appears and is confirmed as slang usage by an LLM judge. Second, we scrape social media platforms and open HuggingFace datasets, applying the same automated validation to confirm both that the sentence contains slang and that the slang term belongs to our curated list. Importantly, all sentences are drawn from naturally occurring human-produced text — no synthetic data was used in constructing the dataset.

The dataset comprises 8,621 labeled instances. The vast majority of literal
counterparts were sourced from the same corpora as the slang samples for each
language, maintaining stylistic consistency within each language.

\paragraph{Validation of Automated Slang/Literal Labels}
\label{app:label_validation}
Portions of the dataset arrive with pre-existing labels, while the remainder were labeled by GPT-4o-mini. To validate the
latter, we manually annotated a balanced sample per language and evaluated the
automated labels against it. In English ($n{=}50$; 25 slang / 25 literal) and
Russian ($n{=}50$; 25 slang / 25 literal), GPT-4o-mini achieves precision $0.96$
and recall $1.00$ on the slang class, with a single literal sentence
misclassified as slang and all slang sentences classified correctly. In Hebrew
($n{=}100$; 50 slang / 50 literal), it reaches precision $0.92$ and recall
$0.98$. These results indicate that the slang/literal distinction is robust and
reliably identifiable by the model. Moreover, our contrastive setup---which
probes the activation delta between slang and literal senses of the same
word---targets a dominant, recurrent latent direction and is therefore stable
under minor label noise, as the aggregate signal remains anchored by the
consistent majority of correctly identified pragmatic contrasts.

\paragraph{Lexical Diversity.} To assess whether slang instances are dominated
by a small subset of target terms, we compute the proportion of slang instances
accounted for by the top-5 most frequent target words per language. The top-5
types account for 22.6\% of English slang instances (219/968, out of 130 target
terms), 58\% of Russian slang instances (312/538, 15 terms), and 78.5\% of Hebrew
slang instances (1,400/1,776, 18 terms), reflecting the larger and more varied
English target-term inventory relative to Hebrew and Russian. 
Top ranked target-terms for all three languages are provided in
Table~\ref{tab:top5_terms}.

\begin{table}[H]
\centering
\small
\begin{tabular}{llll}
\toprule
  & \textbf{English} & \textbf{Hebrew} & \textbf{Russian} \\
\midrule
1 & off (5.7\%)   & {\hebrew אש} (26.6\%)   & {\russian бомба} (12.6\%) \\
2 & man (5.7\%)   & {\hebrew מת} (19\%)   & {\russian жесть} (11.9\%) \\
3 & hit (4.6\%)   & {\hebrew דפוק} (14.1\%) & {\russian пушка} (11.9\%) \\
4 & check (3.7\%) & {\hebrew רצח} (11.1\%)  & {\russian треш} (11.0\%)  \\
5 & bad (2.9\%)   & {\hebrew פצצה} (07.7\%) & {\russian огонь} (10.6\%) \\
\bottomrule
\end{tabular}
\caption{Top-5 most frequent slang target terms per language, with each term's share of that language's total slang instances.}
\label{tab:top5_terms}
\end{table}

\section{Top Discriminative SAE Features by Language}
\label{app:top_features}

This section lists the cross-linguistic and per-language discriminative features
identified in Section~\ref{sec:cross_linguistic_overlap}. Cross-linguistic
features appear in the top-100 discriminative set for all three source
languages; per-language features are each language's top-10 ranked by $\Delta_i$
(slang minus literal activation rate). The two layers share no cross-linguistic
features in common.

\begin{table}[H]
\centering
\small
\begin{tabular}{rcc}
\toprule
\textbf{Rank} & \textbf{Feature} & \textbf{Avg $\Delta_i$} \\
\midrule
1  & 35440 & +33.6\% \\
2  & 93521 & +31.6\% \\
3  & 97003 & +21.2\% \\
4  & 33236 & +19.9\% \\
5  & 88782 & +14.8\% \\
6  & 92777 & +11.0\% \\
7  & 20069 & +10.6\% \\
8  & 71479 & +10.1\% \\
9  & 24006 &  +9.6\% \\
10 & 46374 &  +9.2\% \\
\bottomrule
\end{tabular}
\caption{Cross-linguistic features (Layer~20), ranked by average $\Delta_i$.}
\label{tab:top10_universal_l20}
\end{table}

\begin{table}[H]
\centering
\small
\begin{tabular}{rcccc}
\toprule
\textbf{Rank} & \textbf{Feature} & \textbf{Slang \%} & \textbf{Literal \%} & \textbf{$\Delta_i$} \\
\midrule
\multicolumn{5}{l}{\textit{English}} \\
1  & 35440  & 59.9 & 7.8  & +52.1 \\
2  & 33236  & 45.9 & 2.1  & +43.7 \\
3  & 93521  & 40.7 & 1.0  & +39.7 \\
4  & 7697   & 36.0 & 17.3 & +18.7 \\
5  & 108864 & 25.8 & 7.2  & +18.6 \\
6  & 21876  & 20.1 & 2.4  & +17.7 \\
7  & 28165  & 40.4 & 23.1 & +17.3 \\
8  & 40141  & 42.0 & 25.0 & +17.0 \\
9  & 88782  & 17.3 & 0.3  & +17.0 \\
10 & 103326 & 18.6 & 2.4  & +16.3 \\
\midrule
\multicolumn{5}{l}{\textit{Hebrew}} \\
1  & 115163 & 51.7 & 29.6 & +22.1 \\
2  & 35440  & 24.0 & 2.8  & +21.2 \\
3  & 74327  & 71.1 & 56.6 & +14.5 \\
4  & 93521  & 13.4 & 0.3  & +13.0 \\
5  & 97003  & 16.7 & 3.9  & +12.8 \\
6  & 30218  & 45.6 & 33.2 & +12.4 \\
7  & 7163   & 13.1 & 0.7  & +12.4 \\
8  & 8408   & 17.0 & 5.7  & +11.3 \\
9  & 104622 & 23.5 & 13.0 & +10.5 \\
10 & 86719  & 48.0 & 38.0 & +10.0 \\
\midrule
\multicolumn{5}{l}{\textit{Russian}} \\
1  & 93521  & 43.0 & 1.0  & +42.0 \\
2  & 97003  & 39.8 & 1.7  & +38.1 \\
3  & 115163 & 46.8 & 8.9  & +37.9 \\
4  & 45410  & 37.3 & 7.8  & +29.5 \\
5  & 35440  & 28.8 & 1.3  & +27.5 \\
6  & 83575  & 23.8 & 2.4  & +21.4 \\
7  & 92777  & 21.3 & 0.3  & +21.0 \\
8  & 88782  & 21.5 & 0.6  & +20.9 \\
9  & 24006  & 27.5 & 8.4  & +19.1 \\
10 & 3598   & 21.1 & 2.2  & +19.0 \\
\bottomrule
\end{tabular}
\caption{Top 10 discriminative features per language (Layer~20).}
\label{tab:top10_perlang_l20}
\end{table}

\begin{table}[H]
\centering
\small
\begin{tabular}{rcc}
\toprule
\textbf{Rank} & \textbf{Feature} & \textbf{Avg $\Delta_i$} \\
\midrule
1 & 38266  & +32.5\% \\
2 & 58884  & +32.1\% \\
3 & 13485  & +22.5\% \\
4 & 22467  & +22.4\% \\
5 & 47673  & +18.3\% \\
6 & 106322 & +17.5\% \\
7 & 31252  & +15.2\% \\
8 & 105792 & +15.0\% \\
9 & 88558  & +14.0\% \\
\bottomrule
\end{tabular}
\caption{Cross-linguistic features (Layer~9), ranked by average $\Delta_i$.}
\label{tab:top10_universal_l9}
\end{table}

\begin{table}[H]
\centering
\small
\begin{tabular}{rcccc}
\toprule
\textbf{Rank} & \textbf{Feature} & \textbf{Slang \%} & \textbf{Literal \%} & \textbf{$\Delta_i$} \\
\midrule
\multicolumn{5}{l}{\textit{English}} \\
1  & 58884  & 38.7 & 2.0  & +36.7 \\
2  & 31547  & 39.9 & 4.8  & +35.1 \\
3  & 13485  & 38.8 & 3.7  & +35.1 \\
4  & 38266  & 38.1 & 3.7  & +34.3 \\
5  & 82164  & 38.6 & 4.3  & +34.3 \\
6  & 65823  & 31.0 & 1.1  & +29.9 \\
7  & 99647  & 33.1 & 3.9  & +29.2 \\
8  & 22467  & 31.6 & 4.0  & +27.6 \\
9  & 127209 & 53.8 & 26.8 & +27.1 \\
10 & 124105 & 27.5 & 0.6  & +26.9 \\
\midrule
\multicolumn{5}{l}{\textit{Hebrew}} \\
1  & 38266  & 29.6 & 7.9  & +21.7 \\
2  & 119607 & 76.1 & 54.7 & +21.4 \\
3  & 22467  & 22.8 & 2.8  & +20.1 \\
4  & 64480  & 69.0 & 49.9 & +19.1 \\
5  & 58884  & 29.2 & 10.6 & +18.5 \\
6  & 8819   & 41.0 & 23.2 & +17.8 \\
7  & 13273  & 22.0 & 5.1  & +16.9 \\
8  & 86686  & 31.6 & 15.1 & +16.5 \\
9  & 25081  & 16.9 & 0.4  & +16.5 \\
10 & 39125  & 49.1 & 33.0 & +16.0 \\
\midrule
\multicolumn{5}{l}{\textit{Russian}} \\
1  & 101504 & 57.8 & 15.2 & +42.6 \\
2  & 38266  & 49.1 & 7.7  & +41.5 \\
3  & 58884  & 52.8 & 11.7 & +41.1 \\
4  & 1855   & 49.8 & 10.4 & +39.4 \\
5  & 120652 & 36.3 & 1.1  & +35.1 \\
6  & 106322 & 35.6 & 2.7  & +32.9 \\
7  & 129047 & 36.4 & 5.0  & +31.3 \\
8  & 47673  & 36.8 & 5.8  & +30.9 \\
9  & 60050  & 41.8 & 12.9 & +28.8 \\
10 & 12375  & 39.3 & 12.4 & +27.0 \\
\bottomrule
\end{tabular}
\caption{Top 10 discriminative features per language (Layer~9).}
\label{tab:top10_perlang_l9}
\end{table}





\section{Vocabulary Projection: Top Promoted Tokens per Cross-Linguistic Feature}
\label{app:vocab_proj}
Table~\ref{tab:vocab_proj} lists the top promoted vocabulary tokens for the two highest-$\Delta_i$ cross-linguistic features (35440 and 93521), as well as for their averaged vector. 
Following \citet{gur-arieh-etal-2025-enhancing}, we pass each SAE decoder weight vector through the model's final layer norm and project it through the unembedding matrix; tokens are ranked by projection weight. 
The recurrence of informal expressions such as \textit{dudes}, \textit{vibes}, \textit{swag}, and \textit{funky} across the individual features and their average provides qualitative lexical support for interpreting these directions
as associated with informal register.

\begin{table}[h]
\centering
\small
\begin{tabular}{clll}
\toprule
\textbf{Rank} & \textbf{Feat. 35440} & \textbf{Feat. 93521} & \textbf{Avg (35440+93521)} \\
\midrule
1  & dudes      & emojis     & dudes      \\
2  & guys       & vibes      & vibes      \\
3  & funky      & swag       & badass     \\
4  & fellas     & kaarangay  & funky      \\
5  & dude       & coolest    & guys       \\
6  & homie      & dope       & vibe       \\
7  & rappers    & vibe       & dude       \\
8  & badass     & awesome    & swag       \\
9  & streetwear & swagger    & swagger    \\
10 & dope       & vibing     & coolest    \\
\bottomrule
\end{tabular}
\caption{Top 10 vocabulary tokens promoted by the two highest-$\Delta_i$
cross-linguistic features (35440, 93521) and their averaged vector. All promoted
tokens reflect informal register semantics consistently across individual
features and their combination.}
\label{tab:vocab_proj}
\end{table}

This English-dominant pattern is not confined to the cross-linguistic core.
Feature~115163, which is discriminative in Hebrew ($\Delta_i = +22.1\%$) and
Russian ($\Delta_i = +37.9\%$) but carries essentially no slang--literal signal
in English ($\Delta_i = -0.0\%$), promotes \textit{stuff}, \textit{stupid},
\textit{crazy}, and \textit{gonna}. The Russian-specific Feature~3598
($\Delta_i = +19.0\%$ in Russian; outside the top-100 for both English and
Hebrew) likewise promotes \textit{amazing}, \textit{fantastic}, \textit{good},
and \textit{VERY} among its highest-ranked interpretable tokens. That features
carrying no English discriminative signal still read out through English
vocabulary points to a systematic English bias in the model's unembedding
matrix, rather than to any limitation of the cross-linguistic features.

\section{GPT-4o Formality Evaluation Prompt}
\label{app:formality_prompt}

We use GPT-4o-mini as an automated formality judge to evaluate steered model outputs. Texts are scored in batches of 20 with temperature 0 for reproducibility. The system and user prompts are as follows:

\paragraph{System Prompt.}
\begin{quote}
\small
\texttt{You are a formality scorer. Return only a JSON array of numbers between 0 and 1.}
\end{quote}

\paragraph{User Prompt.}
\begin{quote}
\small
\texttt{Rate the formality of each text below on a scale from 0 to 1.}\\
\texttt{- 0 = very informal, casual, slang, colloquial}\\
\texttt{- 1 = very formal, professional, academic}\\[0.5em]
\texttt{Return ONLY a JSON array of numbers, nothing else. Example: [0.2, 0.8, 0.5]}\\[0.5em]
\texttt{Texts:}\\
\texttt{1. "<text\_1>"}\\
\texttt{2. "<text\_2>"}\\
\texttt{...}
\end{quote}

Scores are clipped to $[0, 1]$. Failed batches are retried up to 3 times. To validate the automated scores, 30 completions per language (120 total) were independently scored by human annotators, yielding a Pearson correlation of $r = 0.62$ between GPT-4o-mini and human judgments.

\subsection{Per-Language Human--LLM Agreement}
\label{app:perlang_agreement}
Per-language Pearson $r$ between GPT-4o-mini and human ratings is $0.768$
(English), $0.803$ (German), $0.474$ (Hebrew), and $0.368$ (Russian), with
$n{=}30$ per language and a pooled $r$ of $0.626$. Higher-resource Latin-script
languages show strong agreement, while Hebrew and Russian show moderate
agreement, broadly in line with published human--human inter-annotator agreement
on formality ($\approx 0.5$--$0.7$; \citealp{pavlick2016formality}). Mean scores
are well-calibrated across all languages ($|\text{GPT mean} - \text{human mean}|
\leq 0.10$), indicating that the lower correlations reflect unbiased per-item
noise rather than systematic bias. Human inter-annotator agreement on the same
items is $r{=}0.814$ (English) and $r{=}0.615$ (Hebrew), placing GPT-4o-mini's
per-language agreement within the range of human--human variability. Judge noise
is therefore non-trivial but does not threaten the rank-level trends our steering
analysis relies on.

\subsection{Prompt Sensitivity of the Formality Judge}
\label{app:prompt_sensitivity}
To assess sensitivity to the judge prompt, we re-scored the 120 human-rated
completions under two rephrasings (V2: ``evaluate level of formality''; V3:
``score the register, $0={}$colloquial / $1={}$formal written''). Inter-prompt
agreement is high (V1--V2 $r{=}0.771$, V1--V3 $r{=}0.773$, V2--V3 $r{=}0.944$;
all $p < 10^{-24}$); per-language correlations span $0.69$--$0.99$ across all
12 prompt$\times$language cells, and 89\% of texts vary by less than $0.20$ on
the $0$--$1$ scale across prompts. 
Prompt reformulation thus introduces minor calibration shifts but reliably preserves the underlying linear relationships. 
These results indicate that the steering evaluation is relatively insensitive to modest reformulations of the judge prompt.

\section{Steering Evaluation Prompts}
\label{app:steering_prompts}

Table~\ref{tab:english_prompts} lists the English sentence prompts used for activation steering evaluation. Each prompt is an incomplete sentence that the model completes under varying steering coefficients. Prompts include both neutral stems (1--15) and informally-primed stems (16--20) to test steering effectiveness across starting registers. Hebrew, Russian, and German were translated by native speakers, and the remaining five languages by GPT-4o; each language has 20 prompts. Full lists are available in the supplementary code.

\begin{table}[H]
\centering
\small
\begin{tabular}{cl}
\toprule
\textbf{\#} & \textbf{Prompt} \\
\midrule
1  & I think that \\
2  & She believes the new restaurant downtown \\
3  & In my opinion, the movie was \\
4  & They feel like the music \\
5  & The exam results were \\
6  & The presentation was \\
7  & The game last night was \\
8  & The outfit she wore looked \\
9  & The book I finished was \\
10 & The car he bought seems \\
11 & I feel that the concert was \\
12 & The coffee from that shop tastes \\
13 & The new phone features are \\
14 & The vacation we took was \\
15 & He thought his explanation was \\
\midrule
16 & Yo, that party was \\
17 & Bro, check out this \\
18 & That beat is so \\
19 & Dude, the game last night was \\
20 & Man, this food is \\
\bottomrule
\end{tabular}
\caption{English steering prompts. Prompts 1--15 are register-neutral; 16--20 are informally primed. Hebrew, Russian, and German prompts follow the same design.}
\label{tab:english_prompts}
\end{table}

\section{Random Feature Steering Ablation}
\label{app:random_ablation}
To rule out the possibility that arbitrary directions in the SAE latent space produce comparable formality shifts---or that steering-induced output degradation is misclassified as informality by the automated judge---we conducted a random feature ablation. We sampled 5 independent sets of 10 random features from the full 131,072-feature dictionary (matching the size of the cross-linguistic feature set), constructed normalized steering vectors by averaging their decoder weights (identical to Eq.~\ref{eq:steering_vector}), and applied them to English and German (zero-shot) at $\alpha \in \{-150, 0, 150\}$. Note that the main experiments use $\alpha \in [-150, 100]$ as the coherence-safe operating range (Appendix~\ref{app:perplexity}); we deliberately extend to $\alpha = 150$ here to stress-test both the random vectors and the automated judge under conditions at the boundary of coherent generation. Each condition produced 20 completions, scored by GPT-4o-mini using the same protocol described in Appendix~\ref{app:formality_prompt}.

As shown in Table~\ref{tab:random_ablation}, cross-linguistic universal features produce strong negative correlations between $\alpha$ and formality (English: $r = -0.791$, $p < 10^{-13}$; German: $r = -0.743$, $p < 10^{-10}$), while random feature vectors yield near-zero correlations (English: mean $r = 0.006 \pm 0.133$; German: mean $r = -0.060 \pm 0.077$). One-sample $t$-tests comparing $|r_{\text{random}}|$ against $|r_{\text{universal}}|$ confirm the difference is highly significant (English: $t = -35.58$, $p < 10^{-5}$; German: $t = -26.41$, $p < 10^{-4}$).

Crucially, this ablation serves a dual purpose. First, it establishes that the formality modulation produced by cross-linguistic features is specific to the identified feature subspace rather than a generic property of high-magnitude residual-stream perturbations: mean formality under random steering remains flat across the full range (English: $0.434 \rightarrow 0.419 \rightarrow 0.435$; German: $0.470 \rightarrow 0.488 \rightarrow 0.439$), whereas the universal vector drives it from $0.615$ to $0.165$ in English and from $0.660$ to $0.205$ in German. Second, it validates the automated judge: since random steering at $\alpha = \pm 150$---including beyond the main operating range---produces coherent completions that receive baseline formality scores ($\sim$0.4--0.5), the low formality scores assigned to universal-feature-steered outputs reflect genuine register shifts rather than an artifact of the judge equating degraded text with informality.

\begin{table}[H]
\centering
\small\resizebox{\columnwidth}{!}{
\begin{tabular}{llcc}
\toprule
\textbf{Language} & \textbf{Vector Type} & \textbf{Pearson $r$} & \textbf{$p$-value} \\
\midrule
English & Universal          & $-0.791$ & $5.6 \times 10^{-14}$ \\
English & Random (mean $\pm$ std) & $0.006 \pm 0.133$ & --- \\
\midrule
German  & Universal          & $-0.743$ & $1.1 \times 10^{-11}$ \\
German  & Random (mean $\pm$ std) & $-0.060 \pm 0.077$ & --- \\
\bottomrule
\end{tabular}}
\caption{Pearson correlations between steering coefficient $\alpha$ and formality score for universal vs.\ random feature vectors (5 independent draws of 10 random features each). Coefficients tested: $\alpha \in \{-150, 0, 150\}$.}
\label{tab:random_ablation}
\end{table}

\section{Language Preservation Under Steering}
\label{app:lang_preservation}

To verify that activation steering modulates formality rather than switching the output language, we applied automatic language identification to all steered completions across nine languages and six coefficients ($\alpha \in \{-150, -100, -50, 0, 50, 100\}$). Overall, approximately 90\% of completions remained in the target language. Preservation was highest at negative (formal-direction) coefficients and decreased at positive coefficients, consistent with the model's English-dominant informal register exerting a pull at high $\alpha$. The observed language switches were directed predominantly toward English, with only isolated switches to other non-target languages. Hebrew, Russian, and Japanese all exceeded 95\% preservation; the remaining languages stayed above 80\%. These results confirm that the formality shifts reported in Section~\ref{sec:causal_experiments} reflect genuine within-language register changes.
We note that approximately 90\% target-language retention is itself a robust result given the difficulty of the setting: steering is applied zero-shot to typologically
distant, low-resource languages (e.g., Amharic, Georgian) never used to derive
the steering features. 
Moreover, the $\sim$10\% of switches are not random but overwhelmingly default to English, consistent with the English bias in the unembedding space noted in \S\ref{sec:vocab_projection}---rather than evidence that the cross-linguistic register features themselves are unstable.

\section{Additional Steering Information}
\label{app:formality_by_language}
\begin{figure}[H]
    \centering
    \includegraphics[width=\columnwidth]{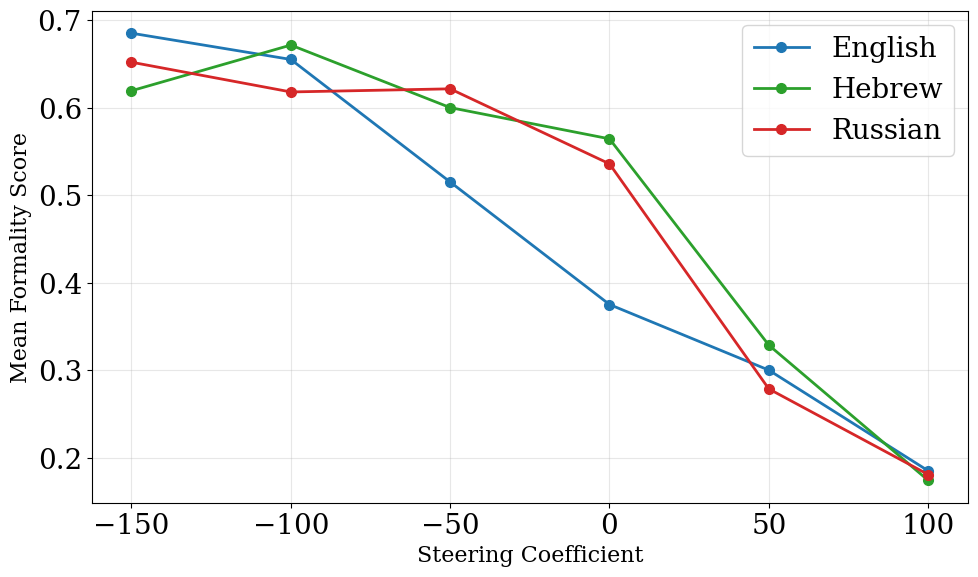}
    \caption{Layer 9: Mean formality score by steering coefficient 
    for each source language.}
    \label{fig:layer9_formality_by_language}
\end{figure}

\begin{figure}[H]
    \centering
    \includegraphics[width=\linewidth]{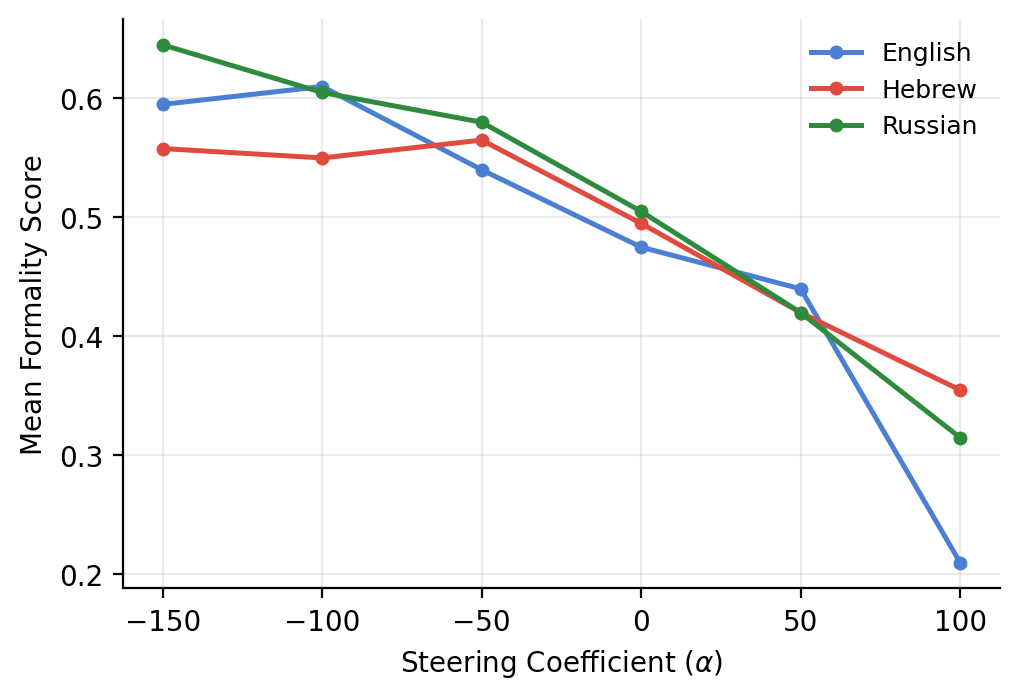}
    \caption{Mean formality score as a function of steering coefficient $\alpha$ 
    for the three source languages (English, Hebrew, Russian) using universal 
    cross-linguistic features at Layer~20.}
    \label{fig:source_lang_steering}
\end{figure}

Table~\ref{tab:layer9_correlations} reports Layer~9 cross-linguistic feature
steering correlations, mirroring the Layer~20 results in the main text, and
Figure~\ref{fig:layer9_formality_by_language} shows the corresponding mean
formality curves by steering coefficient.
Figure~\ref{fig:source_lang_steering} shows the mean formality scores for the
three source languages at Layer~20.
\begin{table}[h]
    \centering
    \begin{tabular}{lcc}
        \toprule
        Language & $r$ & $p$ \\
        \midrule
        English & $-0.623$ & $< 10^{-14}$ \\
        Hebrew  & $-0.637$ & $< 10^{-14}$ \\
        Russian & $-0.631$ & $< 10^{-14}$ \\
        \bottomrule
    \end{tabular}
    \caption{Layer 9: Pearson correlation between steering coefficient 
    and formality score for source languages.}
    \label{tab:layer9_correlations}
\end{table}

\section{Additional Perplexity Analysis}
\label{app:perplexity}
\begin{figure}[H]
    \centering
    \includegraphics[width=\linewidth]{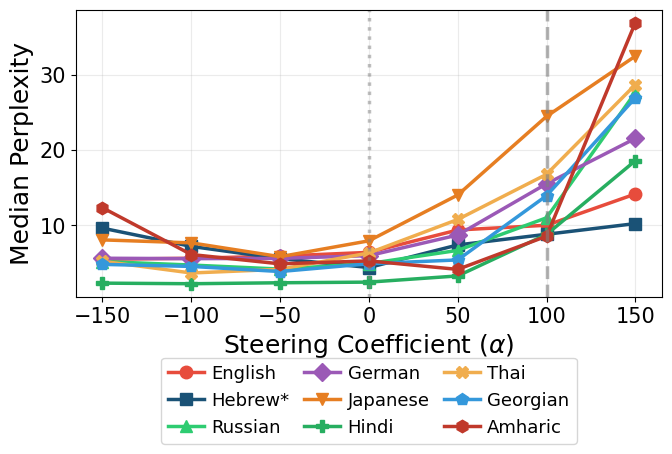}
    \caption{Median perplexity by steering coefficient $\alpha$ for all nine
    languages evaluated with independent reference models (Qwen2.5-7B-Instruct
    for eight languages; DictaLM 2.0 for Hebrew). The stable perplexity within
    $[-150, 100]$ independently corroborates the coherence window identified
    via Gemma self-perplexity (Figure~\ref{fig:perplexity_gemma}).}
    \label{fig:perplexity_external}
\end{figure}

To independently validate that steered outputs remain coherent within our
chosen operating range (Section~\ref{subsec:causal_results},
Figure~\ref{fig:perplexity_gemma}), we additionally compute perplexity using
external reference models: Qwen2.5-7B-Instruct \citep{qwen2.5} for eight
languages and DictaLM 2.0 Instruct \citep{shmidman2024dictalm} for Hebrew, due
to limited Hebrew tokenizer coverage in Qwen2.5. Perplexity is computed on the
completion tokens only, conditioned on the prompt; completions with perplexity
above $10{,}000$ (degenerate or empty generations) are excluded from the
reported medians. As shown in Figure~\ref{fig:perplexity_external}, median
perplexity under these external models remains stable across the full
$[-150, 100]$ range for all nine languages, with values staying below $25$ for
most languages and within $2\times$ of baseline for $|\alpha| \leq 50$. This
cross-model consistency confirms that the formality shifts reported in
Section~\ref{sec:causal_experiments} reflect genuine register changes rather
than degenerate outputs.

\section{Bilingual-Exclusive Feature Steering}
\label{app:bilingual}

As a stricter test of cross-linguistic abstraction, we evaluate
\emph{bilingual-exclusive} feature sets. For each target language, we first
identify features that appear in the top-20 $\Delta_i$ lists of the other two
languages, and then remove any feature that also appears in the target
language's own top-20 list. Thus, when steering English, we use only features
shared by Hebrew and Russian but absent from English's top-20 set; analogous
constructions are used for Hebrew and Russian.

This design makes the transfer test more demanding than the universal-feature
setting: the steering features are not only derived without using the target
language as a source, but are explicitly constrained to exclude the target
language's own most salient slang-associated features. If such feature sets
still support causal control in the held-out language, this strengthens the
claim that the model contains partially language-abstract mechanisms for
informal-register processing, rather than relying solely on language-specific
latent features.

Despite this exclusion, bilingual-exclusive features still yield strong negative
correlations between steering coefficient $\alpha$ and output formality in the
held-out language. 
At Layer~9, the resulting correlations are $r=-0.701$ for English (steered by Hebrew+Russian features), $r=-0.675$ for Hebrew
(English+Russian), and $r=-0.628$ for Russian (Hebrew+English). 
At Layer~20, Hebrew$+$Russian features achieve
$r=-0.724$ for English, English$+$Russian features achieve $r=-0.768$ for Hebrew, and Hebrew$+$English features achieve $r=-0.770$ for Russian, all statistically significant at $p < 10^{-19}$. 
These results provide an additional robustness check on the main transfer findings, showing that causal register control can arise even from feature sets that are shared across
languages but exclude the target language itself---suggesting that informal
register signal extends well beyond the strict trilingual core.